# Embedding of Blink Frequency in Electrooculography Signal using Difference Expansion based Reversible Watermarking Technique

Nilanjan Dey[1] Prasenjit Maji[2] Poulami Das[2] Shouvik Biswas[2] Achintya Das[3]
Sheli Sinha Chaudhuri[1]

**Abstract** – In the past few years, like other fields, rapid expansion of digitization and globalization has influenced the medical field as well. For progress of diagnostic results most of the reputed hospitals and diagnostic centres all over the world have started exchanging medical information. In this proposed method, the calculated diagnostic parametric values of the original Electrooculography (EOG) signal are embedded as a watermark by using Difference Expansion (DE) algorithm based reversible watermarking technique. The extracted watermark provides the required parametric values at the recipient end without any post computation of the recovered EOG signal. By computing the parametric values from the recovered signal, the integrity of the extracted watermark can be validated. The time domain features of EOG signal are calculated for the generation of watermark. In the current work, various features are studied and two major features related to blink frequency are used to generate the watermark. The high Signal to Noise Ratio (SNR) and the Bit Error Rate (BER) claim the robustness of the proposed method.

## I. INTRODUCTION

During the past decade, with the expansion of information digitization and internet, digital media progressively dominated the conventional analog media. As the flip side of the coin, there is an associated disadvantage of digitization and internet. It is also becoming easier for some individuals or groups to copy and transmit digital information without the permission of the owner. The watermark is then introduced to solve this problem. Watermarking is a branch of information hiding which is used to hide proprietary information in digital media like digital images and signals. The ease with which digital content can be exchanged over the internet has created copyright violation issues. Copyrighted material can be easily exchanged over peer-to-peer networks, and this has caused major concerns to those content providers who produce these digital contents. The use of internet is the most important media of globalization. In the present mechanized age, globalization has influenced the medical field also. Intelligent exchange of bio-medical images and bio-signals amongst hospitals and diagnostic centres for mutual availability of therapeutic case studies needs highly efficient and reliable transmission. Watermarking technique is also being used in medical field to protect bio-medical information while transmitting through the wireless media. Watermark is added ownership to increase the level of security and to verify authenticity. Patients' information (Electronic Patient Record), logo of the hospitals or diagnostic centres can be added in the bio-medical data as watermark [1, 2] to prove the intellectual property rights [3]. Addition of watermark in a medical signal or image can cause distortion. As all the bio-medical images and signals convey information required in diagnosis of diseases, any kind of distortion is not acceptable. But for authenticity and security of the information a little amount of distortion can be overlooked. So achieving watermarking technique with lesser amount of distortion in bio-medical data is a challenging task.

EOG stands for Electrooculography. As defined in Stedman's Medical Dictionary, EOG is a method of recording eye position and movements in which electrodes placed on the skin adjacent to the eyes [4] measure changes in standing potential between the front and back of the eyeball as the eyes move. In "eye movement technique," electrodes are placed on either side of the eye to measure the standing potential during the movement of the eye from side to side. The standing potential on the eye is proportional to the potential difference induced between the two electrodes and is recorded.

[1] Electronics and Telecommunications Engineering Dept., Jadavpur University, Kolkata, West Bengal, India.
 e-mail: dey.nilanjan@ymail.com
[2] Computer Science & Engineering Dept., JIS College of Engineering, Kalyani, West Bengal, India.
[3] Electronics & Communication Engineering Dept., Kalyani Govt. Engineering College, Kalyani, India.

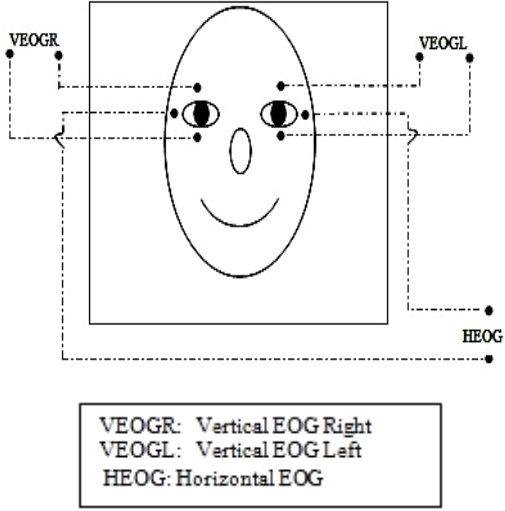

Fig.1. Electrode Placement Scheme

Any disease which affects the rods or the pigment epithelium, gives an abnormal EOG result as in Retinal Detachment, Tapeto-Retinal Degenerations, Vascular Lesions, Myopia, Choroidal Lesions, and Vitamin A Deficiency induced ocular changes, etc [5].

Eye blinking can be measured by EOG. The rate of blinking varies, but on average the eye blinks once every five seconds. A typical blink is about 380 microvolts in amplitude and lasts 120 msec. Eye blinking is a spontaneous process where for a normal adult the blinking rate is 15-20/ min with an interval of 2-10 secs between each blink. Blinking rate can be a criterion for diagnosing diseases. In diseases like People with Parkinson (PWP), the blink occurs less frequently resulting in irritated and dry eyes. On the other hand, excessive blinking may help to diagnose Tourette syndrome, stroke or nervous system disorders, eye twitching and blinking disorder is a symptom of Blepharospasm, etc.

In the first part of this paper, various time domain features are calculated for the generation of watermark. In the second part, the watermark data is hidden within the electrooculogram (EOG) signal using the algorithm Difference Expansion (DE). DE is one of the primitive reversible watermarking techniques. This scheme of watermarking satisfies the following two requirements:

1. Reversible watermarking can retrieve the original information from the watermarked information directly without using the original data [6],

2. Higher Embedding Capacity (The required embedding capacity of the reversible watermarking schemes is much more than the conventional watermarking schemes as this scheme is capable of embedding recovery-information and watermark information in the original information). Finally the Signal to Noise Ratio (SNR) and the Bit Error Rate (BER) is computed to test the overall robustness of the proposed method.

## II. METHODOLOGY

### A. Difference Expansion

In Difference Expansion (DE) method, watermark data can be hidden as well as recovered completely by means of its transform and inverse transform technique [7].

For embedding watermark data within a carrier data, following equations are used:

$$d = m_1 - m_2 \quad (1)$$

$$a = \left\lfloor \frac{m_1 + m_2}{2} \right\rfloor \quad (2)$$

Here, $m_1$ and $m_2$ are integer values of two consequent pixels of the carrier data, $d$ denotes the difference of them and average is denoted by $a$. The value $d$ is then represented into the binary form and again by transforming it into decimal together with the watermark bit $d_w$ is produced.

$$m_1' = a + \left\lfloor \frac{d_w + 1}{2} \right\rfloor \quad (3)$$

$$m_2' = a - \left\lfloor \frac{d_w}{2} \right\rfloor \quad (4)$$

$m_1'$ and $m_2'$ are the new pixel values after adding a watermark bit with the original pixels.

Example: Suppose, $m_1 = 99, m_2 = 93$

Therefore, $d = m_1 - m_2 = 6$ and $a = \left\lfloor \frac{m_1 + m_2}{2} \right\rfloor = 96$

Binary form of d is 110. If the watermark bit B = 1, then by adding it with d, and then converting in decimal $d_w$ is calculated.

Here form of d is 110, therefore 110B = 110B, therefore decimal form $d_w = 13$.

$m_1' = a + \left\lfloor \frac{d_w + 1}{2} \right\rfloor = 103$ and $m_2' = a - \left\lfloor \frac{d_w}{2} \right\rfloor = 90$

At the time of recovering the watermark data from the watermarked data, same equations like the embedding technique are used.

$$d' = m_1' - m_2' \quad (5)$$

$$a' = \left\lfloor \frac{m_1' + m_2'}{2} \right\rfloor \quad (6)$$

where, $d$ is the difference of the watermarked pixel values however $a$ is the average of them. $d$ is then represented into binary form. In the binary data there is only one watermark bit and the remaining bits are for representing the binary form of the pixel value difference of the carrier data. By separating the watermark bit and by converting the remaining bits into decimal the difference *d* between the two original carrier pixels can be recalculated. The original pixel values can be recalculated by using the following two equations:

$$m_1 = a' + \left\lfloor \frac{d+1}{2} \right\rfloor \quad (7)$$

$$m_2 = a' - \left\lfloor \frac{d}{2} \right\rfloor \quad (8)$$

Taking the same example as watermark embedding technique,

$d' = m_1' - m_2' = 13$ and $a' = \left\lfloor \frac{m_1' + m_2'}{2} \right\rfloor = 96$.

Binary form of $d'$ is 1101, now separating the watermark bit B =1 from it and then transforming the remaining bits into decimal d is recalculated.
Here, d = 6. Now the original pixel values can be recalculated,

$m_1 = a' + \left\lfloor \frac{d+1}{2} \right\rfloor = 99$ and $m_2 = a' - \left\lfloor \frac{d}{2} \right\rfloor = 93$.

### B. Feature Analysis of EOG Signal and Watermak Generation

In our proposed method, different features of the original EOG signal are analyzed based on time domain in order to yield a computational simplicity. Computation of time domain features gives a measure of time domain amplitude. In this paper a real time EOG signal obtained from PDS lab [8] as a pilot test is taken. The useful frequency of EOG signal lies within 5 Hz.

Different features of the EOG signal along with the corresponding values obtained from the original EOG signal are described below.

#### 1. Mean absolute value (MAV)

In mathematics, absolute value of a number A means distance between the number and 0 on the number line. As distance can never be negative therefore absolute value of a number can never be negative. Mean absolute value of a set of numbers is calculated by dividing the summation of the absolute values of all the numbers in the set by the total count of the numbers in the set. Mean absolute value of a signal *MAVp* can be calculated by adding the absolute value of all the values in a segment p and dividing it by the length N of the segment. MAV [9] can be represented by the following equation:

$$MAV_p = \frac{1}{N} \sum_{i}^{N} |X_i| \quad (9)$$

#### 2. Standard Deviation

Standard deviation of a random variable is simply the square root of its variance. Standard deviation actually shows the dispersion of the value of a variable from its expected, mean, or average value. Standard deviation can be represented by the following equation:

$$SD = \sqrt{\frac{1}{N} \sum_{i=0}^{N} (x_i - \bar{x})} \quad (10)$$

Table 1

| Input EOG Signal in Volts | |
|---|---|
| | Original EOG |
| Mean Value | 36.1163 |
| Standard deviation | 9.8547 |

After converting the sample EOG signal into micro-volts, data with sample rate 250 and waveform period 130 (Baseline Drift EOG), analysis of following features is done.

#### 3. Maximum Peak Amplitude Value (PAV)

The measure of signal amplitude value at the highest point is maximum peak amplitude value [10].

#### 4. Maximum Valley Amplitude Value (VAV)

The measure of signal amplitude at the lowest point is called maximum valley amplitude value.

#### 5. Maximum Peak Amplitude Position Value (PAP)

It determines the value of amplitude position of the signal at highest point.

#### 6. Maximum Valley Amplitude Position Value (VAP)

It determines the value of amplitude position of the signal at the lowest point.

#### 7. Variance (VAR)

Variance is the average of the squared differences from the mean. It is actually a measure of the signal power. Following equation represents the variance:

$$VAR = \frac{1}{N} \sum_{i=1}^{N} (x_i - \bar{x})^2 \quad (11)$$

#### 8. Areas under curve cycle (AUC)

It determines the summation of absolute value of the signal amplitude under both positive as well as

negative curves. It can be represented by the following equation:

$$AUC = \sum_{i=1}^{N} |x_i| \quad (12)$$

Table 2

|  | Original EOG |
|---|---|
| Maximum Peak Amplitude Value (PAV) | 4.9698 x $10^{-5}$ |
| Maximum valley amplitude value (VAV) | 353 |
| Maximum peak amplitude position value (PAP) | -3.7076 x $10^{-5}$ |
| Maximum valley amplitude position value (VAP) | 53 |
| Areas under curve value (AUC) | 0.0058 |
| Variance of EOG signal (VAR) | 9.6983 x $10^{-11}$ |

*9. Blink Frequency and Blink Interval*

The blink frequency for each blink is calculated and then the amount of blinks exceeding the boundary is detected, which is based on mean value and standard deviation of the blink frequency at the onset of the alert condition. The blink frequency for one blink is calculated as *fi=1/Ti*, where Ti is the time interval between two blinks. The mean value of the blink frequency is calculated as a mean value of all frequencies within the time interval [11]:

$$\overline{f} = \frac{\sum_{i=1}^{n} f_i}{n} \quad (13)$$

The actual number of blinks per time interval is not depicted by this definition of mean value, unless the blink intervals were equally spaced in time. To reveal the exact number of blinks per time interval, the blink frequency has to be termed as the number of blinks in a time interval, i.e. the amount of blink intervals plus one, divided with the length of the time interval:

$$\frac{n+1}{\sum_{i=1}^{n} T_i} \quad (14)$$

Table 3

|  | Original EOG |
|---|---|
| Mean Value of Blink Frequency | 0.3911 |
| Mean Value of Blink Interval | 0.3730 |

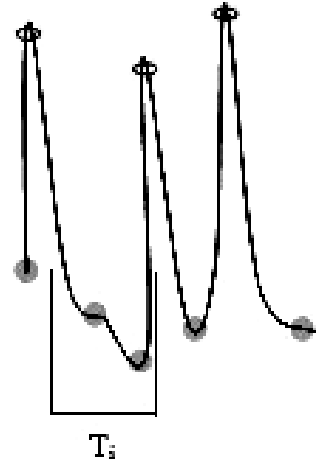

Fig.2. Blink Frequency and Blink Interval

In the current study, Mean Value of Blink Frequency and Mean Value of Blink Interval obtained from the original signal is converted into corresponding binary values (IEEE754 Single precision 32-bit). These 32 bit values are then combined to generate a 64 bit watermark sequence.

Table 4

| Mean Value of Blink Frequency | Mean Value of Blink Interval |
|---|---|
| 0.3911 | 0.3730 |
| 32 bit Binary representation of Mean Value of Blink Frequency | 32 bit Binary representation of Mean Value of Blink Interval |
| 00111110 11001000 00111110 01000010 | *00111110 10111110 11111001 11011011* |
| Generated Watermark(64bit) | |
| [00111110 11001000 00111110 01000010 *00111110 10111110 11111001 11011011*] | |

## III. PROPOSED METHOD

### A. Watermark Embedding

Step 1. The original EOG signal is cropped to a length of 128.
Step 2. Mean Value of Blink Frequency and Mean Value of Blink Interval obtained from the original signal are converted into a 64 bit binary watermark.
Step 3. Difference Expansion algorithm for embedding is applied on the cropped EOG signal to embed the watermark.
Step 4. The rest of the signal is merged with the watermarked EOG signal.

### B. Watermark Extraction

Step 1. At the recipient end, a length of 128 is cropped from the received EOG signal.
Step 2. Difference Expansion algorithm for extraction is applied for recovering the watermark.

## IV. RESULT AND DISCUSSIONS

MATLAB 7.0.1 Software is extensively used for the study of EOG Signal. Concerned images obtained in the result are shown in Fig. 3 through 4.

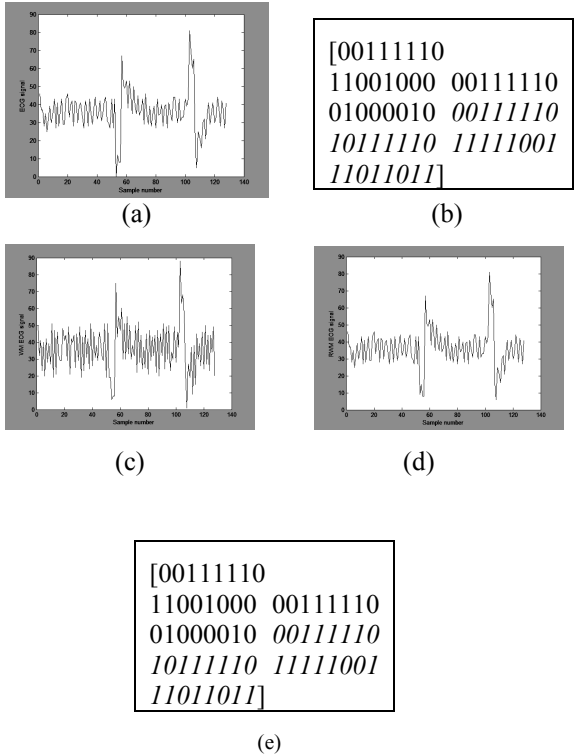

Fig. 3. (a) Original EOG Signal, (b) Watermark, (c) Watermarked EOG Signal, (d) Recovered EOG Signal (e) Recovered Watermark.

### A. Signals to Noise Ratio (SNR)

Signal to Noise Ratio defines the ratio of the signal power to the noise that causes distortion in the signal. High SNR guarantees clear acquisition of the signal with small amount of distortion. A ratio more than 1:1 signifies more signal than the noise [12].

SNR can be represented using the following equation:

SNR = $P_{sig}/P_{no}$

SNR between the original signal and watermarked signal is 33.0385

### B. Bit Error Rate (BER)

The robustness of watermarking algorithm can be measured by a factor namely, Bit Error Rate (BER).

BER = b'/b

Where, b' is the number of incorrectly decoded bits and b is the total number of bits of the original watermark sequence. A BER of 0 signifies that each bit was correctly decoded.

BER between Watermark Bit Sequence and Extracted Watermark Bit Sequence is 0.

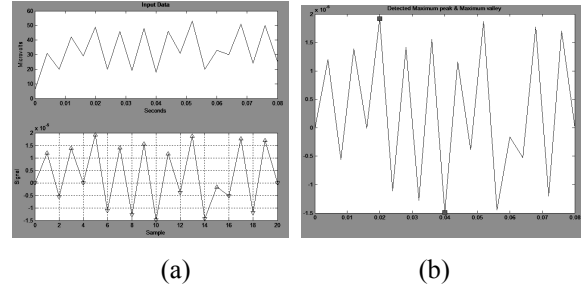

Fig.4. (a) Baseline drifted Watermarked Signal, (b) Maximum Peak and Maximum Valley of Watermarked Signal

## V. CONCLUSION

Since the application of watermarking in Electrooculogram signal is a new field of research, many methodological aspects are used to insert and extract watermarks to and from a signal respectively. Imperative clinical utility is drawn from the innovative application of watermarking technique in electrooculography. Proposed method based on Difference Expansion for Watermark Insertion and Extraction Technique to insert watermark in an EOG signal and to extract watermark from the EOG signal is useful for telemedicine application towards the reduction of diagnostic parameter computation and cross validation of those parameters. After extracting the watermark from the watermarked signal it can be seen that the SNR of the original EOG signal vs. recovered EOG signal is high, which claims the robustness of the proposed method. The BER between the watermark and the extracted watermark bit sequence is 0, which implies that each bit was correctly decoded. Reversible watermarking techniques like Allatar's method [13], Prediction Error [14], Low-Distortion Prediction Error [15] etc. were done by the author earlier with different biosignals. For further extension, this current method may be compared with other existing algorithms in order to draw a distinct conclusion.